\documentclass[11pt, a4paper]{lumia}

\usepackage[sort&compress]{natbib}
\usepackage{xspace}

\theoremstyle{plain}
\newtheorem*{proposition*}{Proposition}

\theoremstyle{definition}

\theoremstyle{definition}

\def\eqref#1{equation~\ref{#1}}
\usepackage{graphicx}           % graphics support
\usepackage{tikz}               % TikZ graphics
\usepackage[edges]{forest}      % forest trees

\usepackage{url}                % simple URL typesetting
\usepackage{xurl}               % extended URL breaking

\usepackage{algorithmic}

\definecolor{mygray}{gray}{0.9}

\usepackage{array}              % extended array and tabular
\usepackage{longtable}          % multi-page tables
\usepackage{multirow}           % multirow cells
\usepackage{makecell}           % enhanced cell formatting
\usepackage{ragged2e}           % ragged text alignment

\usepackage{mathtools}          % additional math tools
\usepackage{nicefrac}           % compact symbols for 1/2, etc.

\usepackage{algorithm}          % algorithm environment
\usepackage{listings}           % code listings

\usepackage{subcaption}         % subfigures and subcaptions
\usepackage{wrapfig}            % text wrapping around figures
\usepackage[export]{adjustbox}  % adjustable boxes

\usepackage[commandnameprefix=ifneeded]{changes}
\usepackage{xspace}             % intelligent spacing
\usepackage[normalem]{ulem}     % underlining without affecting emphasis
\usepackage{CJKutf8}            % CJK character support

\usepackage[tikz]{bclogo}       % logo boxes
\usepackage[framemethod=tikz]{mdframed} % framed text
\usepackage{lipsum}
\usepackage{lipsum}             % lorem ipsum text
\usepackage{tocloft}            % table of contents formatting
\usepackage{afterpage}          % page break control
\usepackage{bbding}             % additional symbols
\usepackage{epigraph}           % epigraphs
\usepackage{minitoc}            % mini table of contents
\usepackage{multicol}           % multiple columns
\usepackage{textgreek}          % Greek text

\newcolumntype{P}[1]{>{\RaggedRight\arraybackslash}p{#1}}

\definecolor{uclablue}{RGB}{39, 116, 174}
\definecolor{bigaired}{RGB}{156, 0, 0}
\definecolor{myblue}{HTML}{598BE7}
\definecolor{mildblue}{RGB}{31,119,180}
\definecolor{sectionblue}{RGB}{70, 130, 180}
\definecolor{methodblue}{RGB}{0, 150, 136}
\definecolor{bgblue}{RGB}{245,243,253}
\definecolor{ttblue}{RGB}{91,194,224}
\definecolor{mygreen}{rgb}{0.64, 0.56, 0.88}
\definecolor{myyellow}{rgb}{0.68, 0.6, 0.1}
\definecolor{fancygreen}{rgb}{0.33, 0.68, 0.20}
\definecolor{salmon}{rgb}{0.94, 0.52, 0.49}
\definecolor{tablegreen}{rgb}{0.82, 0.94, 0.75}
\definecolor{tableblue}{rgb}{0.81, 0.90, 0.94}
\definecolor{tablered}{rgb}{0.97, 0.85, 0.85}
\definecolor{tableorange}{rgb}{0.96, 0.85, 0.81}
\definecolor{myorange}{rgb}{1.0, 0.49, 0.0}
\definecolor{tlgreen}{rgb}{0.33, 0.68, 0.20}
\definecolor{darkgreen}{RGB}{0,100,0}
\definecolor{darkred}{RGB}{200, 0, 0}
\definecolor{customyellow}{HTML}{FFFACD}
\definecolor{refinegreen}{RGB}{0, 128, 75}
\definecolor{scoregreen}{RGB}{34, 139, 34}
\definecolor{hidden-blue}{RGB}{194,232,247}
\definecolor{hidden-black}{RGB}{20,68,106}
\definecolor{yes}{HTML}{C6EFCE}
\definecolor{no}{HTML}{FFC7CE}
\definecolor{partial}{HTML}{FFEB9C}
\definecolor{external}{HTML}{D9E1F2}
\definecolor{hdr}{HTML}{F2F2F2}
\definecolor{GRPOrow}{gray}{0.96}
\definecolor{FlowRLrow}{RGB}{225,236,255}
\definecolor{FlowBlue}{RGB}{80,120,210}
\definecolor{GRPOGray}{gray}{0.35}

\hypersetup{
    colorlinks=true, 
    citecolor=uclablue, 
    linkcolor=bigaired,
    urlcolor=darkblue
}

\setlist[itemize]{leftmargin=20pt, noitemsep, topsep=0pt}

\NewDocumentCommand{\kaiyan}{mO{}}{\textcolor{purple}{\textsuperscript{\textit{kaiyan}}\textsf{\textbf{\small[#1]}}}}
\NewDocumentCommand{\yuxin}{mO{}}{\textcolor{cyan}{\textsuperscript{\textit{yuxin}}\textsf{\textbf{\small[#1]}}}}
\NewDocumentCommand{\bx}{mO{}}{\textcolor{green}{\textsuperscript{\textit{bx}}\textsf{\textbf{\small[#1]}}}}
\NewDocumentCommand{\at}{mO{}}{\textcolor{red}{\textsuperscript{\textit{AT}}\textsf{\textbf{\small[#1]}}}}
\NewDocumentCommand{\re}{mO{}}{\textcolor{blue}{\textsuperscript{\textit{RE}}\textsf{\textbf{\small[#1]}}}}
\NewDocumentCommand{\ybsun}{mO{}}{\textcolor{magenta}{\textsuperscript{\textit{youbang}}\textsf{\textbf{\small[#1]}}}}
\NewDocumentCommand{\runze}{mO{}}{\textcolor{orange}{\textsuperscript{\textit{runze}}\textsf{\textbf{\small[#1]}}}}
\NewDocumentCommand{\add}{mO{}}{\textcolor{darkgreen}{\textsuperscript{\textit{Maybe Consider Discuss}}\textsf{\textbf{[#1]}}}}

\newcommand{\cmark}{\textcolor{darkgreen}{\boldmath$\checkmark$}}
\newcommand{\xmark}{\textcolor{darkred}{\boldmath$\times$}}

\newenvironment{itemize*}%
 {\leftmargini=10pt\begin{itemize}%
  \setlength{\itemsep}{0pt}%
  \setlength{\parskip}{0pt}%
  }%
 {\end{itemize}}

\newenvironment{enumerate*}%
 {\begin{enumerate}%
  \setlength{\itemsep}{0pt}%
  \setlength{\parskip}{0pt}}%
 {\end{enumerate}}

\newcommand{\cellstatus}[1]{%
  \begingroup
  \StrTrim{#1}[\statusval]%
  \IfStrEq{\statusval}{Yes}{\cellcolor{yes}\cmark}{}%
  \IfStrEq{\statusval}{No}{\cellcolor{no}\xmark}{}%
  \IfBeginWith{\statusval}{Yes (}{\cellcolor{yes}\cmark~\textit{\statusval\unskip}}{}%
  \IfStrEq{\statusval}{Partial}{\cellcolor{partial}\textbf{Partial}}{}%
  \IfStrEq{\statusval}{External}{\cellcolor{external}\textbf{External}}{}%
  \endgroup
}

\newtcolorbox{myboxi}[1][]{
  breakable,
  title=#1,
  colback=red!5,
  colbacktitle=red!5,
  coltitle=black,
  fonttitle=\bfseries,
  bottomrule=0pt,
  toprule=0pt,
  leftrule=2pt,
  rightrule=2pt,
  titlerule=0pt,
  arc=0pt,
  outer arc=0pt,
  colframe=red,
}

\newtcolorbox{myboxnote}[1][]{
  breakable,
  title=#1,
  colback=orange!0,
  colbacktitle=orange!0,
  coltitle=black,
  fonttitle=\bfseries,
  bottomrule=0pt,
  toprule=0pt,
  leftrule=2pt,
  rightrule=2pt,
  titlerule=0pt,
  arc=0pt,
  outer arc=0pt,
  colframe=orange,
}

\newtcolorbox{myboxii}[1][]{
  breakable,
  freelance,
  title=#1,
  colback=white,
  colbacktitle=white,
  coltitle=black,
  fonttitle=\bfseries,
  bottomrule=0pt,
  boxrule=0pt,
  colframe=white,
  overlay unbroken and first={
  \draw[red!75!black,line width=3pt]
    ([xshift=5pt]frame.north west) -- 
    (frame.north west) -- 
    (frame.south west);
  \draw[red!75!black,line width=3pt]
    ([xshift=-5pt]frame.north east) -- 
    (frame.north east) -- 
    (frame.south east);
  },
  overlay unbroken app={
  \draw[red!75!black,line width=3pt,line cap=rect]
    (frame.south west) -- 
    ([xshift=5pt]frame.south west);
  \draw[red!75!black,line width=3pt,line cap=rect]
    (frame.south east) -- 
    ([xshift=-5pt]frame.south east);
  },
  overlay middle and last={
  \draw[red!75!black,line width=3pt]
    (frame.north west) -- 
    (frame.south west);
  \draw[red!75!black,line width=3pt]
    (frame.north east) -- 
    (frame.south east);
  },
  overlay last app={
  \draw[red!75!black,line width=3pt,line cap=rect]
    (frame.south west) --
    ([xshift=5pt]frame.south west);
  \draw[red!75!black,line width=3pt,line cap=rect]
    (frame.south east) --
    ([xshift=-5pt]frame.south east);
  },
}

\tcbset{
  takeawaysbox/.style={
    title=Takeaways,
    colback=lightblue!80,
    colframe=black,
    fonttitle=\bfseries\small,
    coltitle=white,
    colbacktitle=black,
    enhanced,
    attach boxed title to top left={xshift=2.5mm,yshift=-2.5mm},
    boxed title style={rounded corners, size=small, colframe=black, colback=black},
    width=\linewidth,
    arc=3.5mm
  }
}

\mdfdefinestyle{mystyle}{%
  rightline=true,
  innerleftmargin=10,
  innerrightmargin=10,
  outerlinewidth=3pt,
  topline=false,
  rightline=true,
  bottomline=false,
  skipabove=\topsep,
  skipbelow=\topsep
}

\tikzset{%
    every node/.style={font=\tiny},
    parent/.style =          {align=center,text width=2cm,rounded corners=3pt, line width=0.3mm, fill=gray!10,draw=gray!80},
    child/.style =           {align=center,text width=2.0cm,rounded corners=3pt, fill=blue!10,draw=blue!80,line width=0.3mm},
    grandchild/.style =      {align=center,text width=2cm,rounded corners=3pt},
    greatgrandchild/.style = {align=center,text width=1.5cm,rounded corners=3pt},
    greatgrandchild2/.style = {align=center,text width=1.5cm,rounded corners=3pt},    
    referenceblock/.style =  {align=center,text width=1.5cm,rounded corners=2pt},
    pretrain/.style =           {align=center,text width=2.0cm,rounded corners=3pt, fill=blue!10,draw=blue!80,line width=0.3mm},   
    pretrain_work/.style =           {align=center, text width=8.5cm,rounded corners=3pt, fill=blue!10,draw=blue!0,line width=0.3mm},  
    template/.style =           {align=center,text width=2.0cm,rounded corners=3pt, fill=red!10,draw=red!80,line width=0.3mm},   
    template_work/.style =           {align=center,text width=8.5cm,rounded corners=3pt, fill=red!10,draw=red!0,line width=0.3mm},    
    answer/.style =           {align=center,text width=2.0cm,rounded corners=3pt, fill= cyan!10,draw= cyan!80,line width=0.3mm},   
    answer_work/.style =           {align=center,text width=8.5cm,rounded corners=3pt, fill= cyan!10,draw= cyan!0,line width=0.3mm},      
    multiple/.style =           {align=center,text width=2.0cm,rounded corners=3pt, fill= orange!10,draw= orange!80,line width=0.3mm},   
    multiple_work/.style =           {align=center,text width=8.5cm,rounded corners=3pt, fill= orange!10,draw= orange!0,line width=0.3mm},        
    tuning/.style =           {align=center,text width=2.0cm,rounded corners=3pt, fill= magenta!10,draw= magenta!80,line width=0.3mm},   
    tuning_work/.style =           {align=center,text width=8.5cm,rounded corners=3pt, fill= magenta!10,draw= magenta!0,line width=0.3mm},          
}

\newcommand{\lstbg}[3][0pt]{{\fboxsep#1\colorbox{#2}{\strut #3}}}

\lstdefinelanguage{diff}{
  basicstyle=\ttfamily\small,
  morecomment=[f][\lstbg{red!20}]-,
  morecomment=[f][\lstbg{green!20}]+,
}

\lstdefinelanguage{diffpython}{
  language=diff,
  morekeywords={def, if, else, for, while, return, import, from, as, class, with, try, except, finally, raise, lambda, and, or, not, in, is, None, True, False},
  morecomment=[l]{\#},
  morestring=[b]",
  morestring=[b]',
}

\setheadertext{LUMIA Lab}

\correspondingemail{\emailicon~\hyperlink{mailto:songshixiang@sjtu.edu.cn}{songshixiang@sjtu.edu.cn} \quad \hyperlink{mailto:lin.zhouhan@gmail.com}{lin.zhouhan@gmail.com} \quad  $^*$ Equal Contribution. \quad $^\ddagger$ Corresponding Author. }

\renewcommand{\today}{2026-03-02}

\title{AdaPonderLM: Gated Pondering Language Models with Token-Wise Adaptive Depth}
\author{%
  Shixiang Song$^{1,5*}$, He Li$^{1*}$, Zitong Wang$^{1,3}$, Boyi Zeng$^1$, Feichen Song$^1$, Yixuan Wang$^{1,5}$, Zhiqin John Xu$^4$, Ziwei He$^5$, Zhouhan Lin$^{1,2,5\ddagger}$\\
  $^1$ LUMIA Lab, School of Artificial Intelligence, Shanghai Jiao Tong University, $^2$ Shanghai AI Laboratory,\\  $^3$ Sun Yat-sen University, $^4$ Shanghai Jiao Tong University, $^5$ Shanghai Innovation Institute
}

\begin{document}

% ====================
% ABSTRACT
% ====================
\begin{abstract}
% Test-time scaling improves the reasoning ability of large language models by allocating more computation at inference, and its representative, recurrent/iterative Transformers enable test-time scaling directly from pretraining, yet most of them use a fixed number of recurrence steps, wasting compute on easy tokens and lacking token-wise adaptivity. With the core idea of \textit{Adaptive Compuataion Time}(ACT) and \textit{Early Exit}(EE), we propose \textbf{AdaPonderLM}, an early-exit recurrent language model with an end-to-end, self-supervised training framework. We apply a gating mechanism to prune recurrent language model at when to stop, and we introduce a KV-reuse mechaism during training and inference time. We pretrain AdaPonderLM with Pythia 70M and 410M suit, as well as continue pretrain PonderLM with Pythia 1.4B and 2.8B suit and vanilla Pythia 1.4B model. Our results show that AdaPonderLM could significantly reduce the inference FLOPs while maintaining language modeling performance as well as downstream tasks. We also explore how the gating system works during the training and inference time.  

Test-time scaling via recurrent/iterative Transformers enables large language models to spend more computation at inference, but most pretrained recurrent LMs run a fixed number of iterations, wasting compute on easy tokens and lacking token-wise adaptivity. Following the core idea of Adaptive Computation Time(ACT) and Early Exit(EE), we propose \textbf{AdaPonderLM}, a self-supervised recurrent language model that learns \emph{token-wise early exiting} during pretraining without manually tuned per-token/per-layer pruning ratios. AdaPonderLM uses iteration-specific MLP gates with a monotonic halting mask to decide when each token stops recurring, and introduces a KV reuse mechanism that reuses cached key/value states for halted tokens, ensuring train--test consistency and practical acceleration. Across Pythia backbones from 70M to 410M (pretraining) and up to 2.8B (continued pretraining), AdaPonderLM reduces inference compute at about 10\% while maintaining comparable language modeling perplexity and competitive downstream accuracy. Our analysis shows the learned gates allocate more computation to high-NLL (hard) tokens, exhibiting adaptive computation time behavior in a fully self-supervised setting. Meanwhile, under iso-FLOPs, the learned halting policy consistently outperforms fixed pruning, showing AdaPonderLM allocates compute to the right tokens rather than just reducing average depth.

\end{abstract}

\maketitle

\section{Introduction}

\begin{wrapfigure}{r}{0.6\linewidth}
% \begin{figure}[h]
    \centering
    \includegraphics[width=\linewidth]{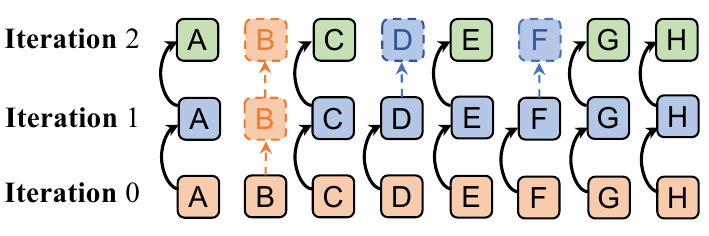}
    \caption{Illustration of the gating behavior at inference time. Across recurrent iterations, tokens that are pruned reuse their KV cache from the previous iteration, avoiding redundant computation. In this example, token \textbf{B} is halted by the gating MLP after the first iteration; its KV cache is then reused in the second and third iterations. Tokens \textbf{D} and \textbf{F} are halted in the second iteration and similarly reuse their KV caches thereafter.}
    \label{fig:inference}
    \vskip -0.2in
% \end{figure}
\end{wrapfigure}

%% 保留
Recently, Test Time Scaling has emerged as a promising direction to improve model performance by allocating additional computation at inference time.
A promising line of research achieves test-time scaling through \emph{recurrent} or \emph{iterative} computation in Transformers. Methods such as Universal Transformer~\citep{dehghani2018universal}, Loop Transformer~\citep{saunshi2025reasoning}, PonderLM~\citep{zeng2025ponderlm, zeng2025pretraining}, Mixture-of-Recursions (MoR)~\citep{bae2025mixture}, and Ouro~\citep{zhu2025scaling} introduce architectures that iteratively refine hidden states or aggregate recurrence via probability-weighted embeddings. These designs effectively extend computation budget at inference time. However, many such recurrent approaches rely on a \emph{fixed} number of iterations in practice, which results in suboptimal resource allocation: wasting computation on trivial tokens while under-serving complex ones.
%A key question is therefore whether we can incorporate the spirit of Adaptive Computation Time (ACT) into recurrent LLM pretraining, so that the model can \emph{adaptively} decide when to stop iterating and thereby save compute during inference.

Following the core idea of Adaptive Computation Time (ACT)~\citep{graves2016adaptive}, we highlight \emph{Early Exit (EE)} as a practical way to alleviate the inefficiency of fixed-depth recurrence. EE has been widely studied in CNNs and Transformers to reduce computation by skipping layers~\citep{teerapittayanon2016branchynet, bolukbasi2017adaptive, xin2020deebert, schwartz2020right}. In decoder-only architectures, methods such as Universal Transformer~\citep{dehghani2018universal} and LayerSkip~\citep{elhoushi2024layerskip} further demonstrate the potential of early exiting.
In recurrent architectures, however, the same principle is particularly natural for enabling \emph{temporal adaptivity}: rather than traversing a predetermined number of recurrent steps, the model can learn to halt its recurrence dynamically. Concretely, specialized halting gates or confidence-based criteria can determine an input-dependent termination time, thereby decoupling computational cost from parameter count and allocating variable amounts of ``thinking time'' to tokens according to their inherent difficulty~\citep{graves2016adaptive,bengio2015conditional}.
%% 加一些Early Exit的方法，Universal Transformer后面一些引用  Early Exit在CNN transformwe
%% 最近有个reburrent model ponderLM, 通过复用参数提升表现，但是因为一些问题导致大家都只能ponder一样步数
%% 对于ponderLM, early exit方法很有用。我们想到一种方法，做work了。。。
%% 

% Existing attempts toward ACT-style recurrent LMs still face important issues. Ouro~\citep{zhu2025scaling} suffers from train--test inconsistency due to mismatched behaviors between pretraining and inference. 
% MoR~\citep{bae2025mixture} enforces train--test consistency, but does not truly reduce computation in practice: its reported comparisons can be unfair when accounting for actual training FLOPs, and its adaptive behavior is limited because pruning lengths are fixed per layer. %%% TODO: 轻喷
% TaH~\citep{fu2025think} achieves dynamic pruning, but it is not a pretraining method and relies entirely on a fully trained teacher model to make decisions, rather than learning the stopping policy end-to-end from self-supervision.

In this work, we propose \textbf{AdaPonderLM}, an early-exit recurrent language model with an end-to-end, self-supervised pre-training framework. AdaPonderLM is built upon \textit{PonderLM}~\citep{zeng2025pretraining}, a recurrent language model balancing training FLOPs and language modeling performance. Unlike fixed-depth recurrence, \textbf{AdaPonderLM} learns to dynamically allocate computation depth at inference time by a gating mechanism without any additional supervised signal, achieving improved trade-offs between modeling quality and inference efficiency.

Our contributions are summarized as follows:
\begin{itemize}
    \item We propose an \textbf{end-to-end pretraining} approach that jointly learns recurrent refinement and stopping behavior in a fully self-supervised manner. With this training approach, we can prune insignificant pondering steps while maintaining language modeling capability.
    \item We propose a model architecture can \textbf{automatically allocate} its pondering budget across tokens \textbf{without manual tuning} of per-token or per-layer pruning ratios.
    \item We introduce a practical \textbf{KV reuse} mechanism for recurrent decoding, enabling \textbf{pruning and acceleration} of Ponder-style recurrence with improved perplexity--compute trade-offs.
    \item We pre-train \textbf{AdaPonderLM} on the Pythia 70M and 410M suites, and further conduct continued pre-training of PonderLM on the Pythia 1.4B and 2.8B suites, as well as a vanilla Pythia-1.4B baseline.
\end{itemize}

\section{Related Work}
\paragraph{Recurrent Language Models.}
The evolution of recurrence in Transformer-based architectures has shifted from parameter efficiency to latent space reasoning. Early paradigms, such as Universal Transformers~\citep{dehghani2018universal}, decoupled depth from parameter count by sharing weights. Subsequent studies~\citep{bai2019deep} generalized this concept by modeling infinite-depth implicit representations through fixed-point solving. In the context of modern LLMs, recurrence is repurposed to scale test-time compute, primarily developing into two mechanisms: (1) Horizontal Recurrence, exemplified by Coconut~\citep{tack2025llm}, which appends continuous latent states to the sequence to form a latent chain-of-thought; and (2) Vertical Recurrence, such as Looped Transformers~\citep{geiping2025scaling,saunshi2025reasoning} and PonderLM~\citep{zeng2025ponderlm,zeng2025pretraining}, which models reasoning as repeated vertical computations within a single generation step.

\paragraph{Adaptive Computation.}
Adaptive computation aims to allocate computational resources dynamically according to the difficulty of the input or individual tokens. Early representative works such as ACT ~\citep{graves2016adaptive} and PonderNet ~\citep{banino2021pondernet} introduce probabilistic halting mechanisms that allow models to learn when to stop computation. In parallel, early-exit mechanisms were developed in CNNs, where lightweight exit heads are attached to intermediate layers and inference is terminated early based on confidence, entropy, or prediction stability, significantly reducing inference cost while preserving performance ~\citep{xin2020deebert,fan2019reducing,zhou2020bert,xie2021elbert,bajpai2025beem}.

After that, early-exit mechanisms were incorporated into Transformer architectures to enable token- or sequence-level adaptive computation  like Universal Transformer~\citep{dehghani2018universal, chen2023ee,shan2024early}, and adjust the number of effective layers dynamically to model reasoning depth~\citep{yang2025specexit, elhoushi2024layerskip, miao2024efficient}. For generative models, adaptivity evolved into explicit sequence extension. Some methods allocates more computation to difficult examples by introducing auxiliary tokens or intermediate reasoning steps ~\citep{goyal2023think,zelikman2024quiet,zelikman2022star}. Another class compresses multi-step reasoning into continuous latent spaces, such as CoCoMix~\citep{tack2025llm} and Token Assorted methods \citep{su2025token,zhang2025soft,tan2025think}, implicitly controlling effective computation depth through latent representations.

However, current frameworks still have notable limitations. Mainstream recurrent models~\citep{zeng2025pretraining} typically assume a fixed compute budget, disregarding token-wise difficulty. Prior adaptive methods often (i) depend on hard supervision and are introduced only after pretraining~\citep{fu2025think}, (ii) are limited to static pruning schemes~\citep{bae2025mixture}, or (iii) require substantial additional data~\citep{zhu2025scaling}. In contrast, we propose a self-supervised adaptive mechanism incorporated into pretraining, eliminating the need for external hard labels.

\section{Methodology}
\begin{figure*}[t]
\centering
\includegraphics[width=\linewidth]{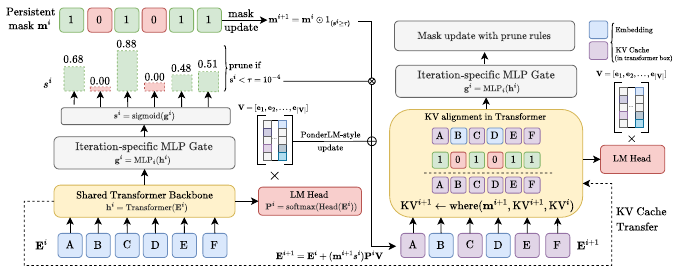}
\caption{Mechanism overview. The same Transformer is executed recurrently across iterations. At iteration $i$, an iteration-specific MLP produces gate probabilities, which update a persistent mask $m^{i}$. KV states are aligned token-wise via \texttt{where}: pruned tokens reuse cached KV states, ensuring training--inference consistency.}
\label{fig:pretrain}
% \vskip -0.2in
\end{figure*}
Our method consists of three parts: the basic model framework, the training-inference consensus mechanism, and the pretraining algorithm. The overview of the mechanism could be seen in~\autoref{fig:pretrain}.
\subsection{MLP-based Gate: Model Framework}
\label{sec:mlp-gate-framework}

\paragraph{Embedding initialization.}
Let the embedding matrix for vocabulary be $V=[e_{1},e_{2},\ldots,e_{|\mathcal{V}|}]\in\mathbb{R}^{|\mathcal{V}|\times d}$.
Given an input token sequence of length $n$, we obtain the initial embeddings
$E^{0}=[e^{0}_{1},e^{0}_{2},\ldots,e^{0}_{n}]$, where $e^{0}_{t}\in\mathbb{R}^{d}$.
We then run a recurrent refinement process indexed by iteration superscripts $i$.

\paragraph{Backbone representation.}
In iteration $i$, the shared Transformer backbone computes:
\begin{equation}
h^{i}=\mathsf{Transformer}(E^{i}).
\label{eq:backbone}
\end{equation}

\paragraph{MLP gate (per-iteration).}
We use a two-layer MLP to produce a token-wise gate logit:
\begin{equation}
g^{i}=\mathrm{MLP}_{i}(h^{i}),
\label{eq:gate_logit}
\end{equation}
and map it to $(0,1)$ by a sigmoid to obtain the gating probability (denoted by $s$):
\begin{equation}
s^{i}=\mathrm{sigmoid}(g^{i}).
\label{eq:gate_prob}
\end{equation}
We empirically observe that sharing a single MLP across all iterations is unstable; therefore, we use an independent $\mathrm{MLP}_i$ for each iteration (with an ablation against the shared-MLP variant).

\paragraph{Pruning rule and persistent mask.}
Following gate-based pruning practice, we prune when the gate is below a threshold $\tau$ (we set $\tau=10^{-4}$):
\begin{equation}
\text{prune at iteration } i \quad \Leftrightarrow \quad s^{i}<\tau.
\label{eq:prune_rule}
\end{equation}
We also maintain a persistent binary mask $m^{i}\in\{0,1\}^{n}$ to enforce monotonic pruning:
\begin{equation}
m^{0}=\mathbf{1},\qquad
m^{i+1}=m^{i}\odot \mathbf{1}(s^{i}\ge \tau),
\label{eq:persistent_mask}
\end{equation}
so once a token is pruned ($m^{i}_{t}=0$), it stays pruned in later iterations.

\paragraph{PonderLM-style probability-weighted update.}
We construct an update embedding using the model's next-token distribution as in PonderLM~\citep{zeng2025pretraining}.
Let the LM head output logits $z^{i}\in\mathbb{R}^{n\times|\mathcal{V}|}$ and probabilities:
\begin{equation}
P^{i}=\mathrm{softmax}(z^{i})\in\mathbb{R}^{n\times|\mathcal{V}|}.
\label{eq:lm_prob}
\end{equation}
The candidate update embedding at position $t$ is the expectation over vocabulary embeddings:
\begin{equation}
e^{i}_{t}=\sum_{v=1}^{|\mathcal{V}|} P^{i}_{t,v}\, e_{v}\in\mathbb{R}^{d},
\label{eq:update_token}
\end{equation}
equivalently in matrix form:
\begin{equation}
E^{i}_{\mathrm{upd}} = P^{i}V\in\mathbb{R}^{n\times d}.
\label{eq:update_matrix}
\end{equation}

\paragraph{Gated embedding update.}
Finally, we update embeddings with a gated residual connection:
\begin{equation}
E^{i+1}_{t}=E^{i}_{t}+\big(m^{i+1}_{t}\, s^{i}_{t}\big)\, e^{i}_{t}.
\label{eq:gated_update}
\end{equation}

We specify our pruning and refinement framework in Alogrithm~\ref{alg:iterative-pruning}.
% \begin{algorithm}[tb]
%   \caption{Bubble Sort}
%   \label{alg:example}
%   \begin{algorithmic}
%     \STATE {\bfseries Input:} data $x_i$, size $m$
%     \REPEAT
%     \STATE Initialize $noChange = true$.
%     \FOR{$i=1$ {\bfseries to} $m-1$}
%     \IF{$x_i > x_{i+1}$}
%     \STATE Swap $x_i$ and $x_{i+1}$
%     \STATE $noChange = false$
%     \ENDIF
%     \ENDFOR
%     \UNTIL{$noChange$ is $true$}
%   \end{algorithmic}
% \end{algorithm}
\subsection{KV Alignment}
\label{sec:kv-alignment}

\begin{wrapfigure}{r}{0.5\linewidth}
\begin{minipage}{\linewidth}
\paragraph{$\quad$}
\begin{algorithm}[H]
\caption{Iterative Pruning with Recurrent Refinement}
\label{alg:iterative-pruning}
\begin{algorithmic}[1]
\REQUIRE Input sequence of length $n$; Vocabulary matrix $V\in\mathbb{R}^{|\mathcal{V}|\times d}$; pruning threshold $\tau$ (e.g., $10^{-4}$); max iterations $K$.
\ENSURE Hidden states $h^{K-1}$.
\STATE Initialize $E^{0}=[e^{0}_{1},\ldots,e^{0}_{n}]$ from $V$
\STATE $m^{0} \gets \mathbf{1} \in \{1\}^{n}$ \COMMENT{Initialize persistent mask to all ones}
\FOR{$i \gets 0$ to $K-1$}
    \STATE $h^{i} \gets \mathsf{Transformer}(E^{i})$
    \STATE $g^{i} \gets \mathrm{MLP}_{i}(h^{i})$ 
    \STATE $s^{i} \gets \mathrm{sigmoid}(g^{i})$ 
    \STATE $m^{i+1} \gets m^{i} \odot \mathbf{1}(s^{i} \ge \tau)$ 
    \STATE $P^i\gets \text{softmax}(\mathtt{lm\_head}(h^i))$
    \STATE $E_\text{upd}^i\gets P^iV$
    \STATE $E^{i+1}\gets E^i+(m^{i+1}s^i)E_\text{upd}^i$
\ENDFOR
\end{algorithmic}
\end{algorithm}  
% \lipsum
\end{minipage}
\vskip -0.3in
\end{wrapfigure}

\paragraph{Motivation.} Our model employs recurrent refinement with token-wise pruning, as outlined in~\autoref{fig:pretrain}. A critical challenge is maintaining consistency between training objectives and inference dynamics. Specifically, once a token is pruned, its representation must remain static across subsequent iterations, as shown in~\autoref{fig:inference}. To achieve this, we implement a KV cache alignment mechanism: for any pruned token, we strictly reuse the Key and Value states from the previous iteration rather than generating new ones.

% \vskip -0.3in
\paragraph{Token-wise KV replacement.}
Let the token-wise persistent mask at iteration $i$ be $m^{i}\in\{0,1\}^{n}$, where $m^{i}_{t}=1$ means token $t$ is still active (not pruned). For a Transformer layer $\ell$, denote the current iteration's key/value by
$K^{i,\ell},V^{i,\ell}\in\mathbb{R}^{B\times H\times n\times d_h}$ and the previous iteration's key/value by
$K^{i-1,\ell},V^{i-1,\ell}$.
We align KV by a masked replacement:
\begin{equation}
\begin{array}{l}
K^{i,\ell} \leftarrow \mathrm{where}(m^{i},\, K^{i,\ell},\, K^{i-1,\ell}),\\
V^{i,\ell} \leftarrow \mathrm{where}(m^{i},\, V^{i,\ell},\, V^{i-1,\ell}),
\end{array}
\label{eq:kv_align}
\end{equation}
where $m^{i}$ is broadcast to match the tensor shape. Intuitively, active tokens use newly computed K/V, while pruned tokens keep the previous iteration's K/V. We specifc the algorithm in Alogrithm~\ref{alg:kv-alignment}.

\begin{wrapfigure}{r}{0.5\linewidth}
\begin{minipage}{\linewidth}
\paragraph{$\quad$}
\vskip -0.4in
\begin{algorithm}[H]
\caption{KV Alignment for Training--Inference Consistency}
\label{alg:kv-alignment}
\begin{algorithmic}[1]
\REQUIRE Current caches $\{(K^{i,\ell},V^{i,\ell})\}_{\ell=1}^{L}$; previous caches $\{(K^{i-1,\ell},V^{i-1,\ell})\}_{\ell=1}^{L}$ (may be \texttt{None});
persistent mask $m^{i}\in\{0,1\}^{B\times n}$.
\ENSURE Aligned caches $\{(K^{i,\ell}_{\mathrm{al}},V^{i,\ell}_{\mathrm{al}})\}_{\ell=1}^{L}$.
\STATE Broadcast $m^{i}$ to $M\in\{0,1\}^{B\times 1\times n\times 1}$
\FOR{$\ell \gets 1$ to $L$}
    \STATE $K^{i,\ell}_{\mathrm{al}} \gets \mathrm{where}(M, K^{i,\ell}, K^{i-1,\ell})$
    \STATE $V^{i,\ell}_{\mathrm{al}} \gets \mathrm{where}(M, V^{i,\ell}, V^{i-1,\ell})$
\ENDFOR

\end{algorithmic}

\end{algorithm}
\end{minipage}
\vskip -0.5in
\end{wrapfigure}
% \vskip -0.1in
Since it does not change the attention operator itself , it is straightforward to integrate with optimized attention kernels such as Flash Attention~\citep{dao2022flashattention}, enabling practical training/inference speedups compared to approaches that require irregular control flow or recomputing pruned-token attention states.

% \paragraph{Scheduling the pruning ratio.}
% We apply a scheduler to control the pruning ratio $k$ during training: early steps use a smaller pruning ratio and gradually increase it to the target value, which stabilizes optimization.
\subsection{Pretraining Algorithm}
\label{sec:pretrain}

We adopt a two-stage training strategy to stabilize optimization.

\paragraph{Stage 1 (warm-up).}
We first train the language model without pondering regularization:
\begin{equation}
\mathcal{L} \;=\; \mathcal{L}_{\text{CE}}.
\label{eq:stage1}
\end{equation}

\paragraph{Stage 2 (pondering regularization).}
We then enable pondering regularization:
\begin{equation}
\mathcal{L} \;=\; \mathcal{L}_{\text{CE}} \;+\; \lambda\,\mathcal{L}_{\text{ponder}} .
\label{eq:stage2}
\end{equation}

\paragraph{Ponder loss (bottom-$K$).}
Let $g$ denote the collection of gate values across all tokens and recurrent iterations. We penalize the bottom-$K$ fraction of gate values:
\begin{equation}
\mathcal{L}_{\text{ponder}} \;=\; \mathrm{bottomK}_{k_s}(g),
\label{eq:ponder_loss}
\end{equation}
where $\mathrm{bottomK}_{k_s}(\cdot)$ returns the mean of the smallest $k_s$ fraction of elements.

\paragraph{Warmup on $k$ in Stage 2.}
During Stage 2, we apply a warmup schedule to the fraction $k_s$ (by training step $s$):
\begin{equation}
k_s=
\begin{cases}
% 0, & s \le S_0,\\[4pt]
\dfrac{s-S_0}{S_1-S_0}\cdot k_{\max}, & S_0 < s < S_1,\\[8pt]
k_{\max}, & s \ge S_1,
\end{cases}
\label{eq:k_warmup}
\end{equation}
where $S_0$ is the delay point (the step at which Stage~2 starts), $S_1$ denotes the end of the warmup period, and $k_{\max}$ is the target bottom-$K$ fraction used after warmup.
\section{Experiments}
Our experiments consist of four main parts:
\begin{enumerate}
\item We pretrain AdaPonderLM (Pythia 70M, 410M) and other recurrent model baselines, including Pause Token~\citep{zelikman2024quiet}, Loop Transformer~\citep{saunshi2025reasoning}, and PonderLM~\citep{zeng2025ponderlm}. We find that AdaPonderLM achieves comparable evaluation loss to PonderLM while reducing inference steps by approximately 8--10\%.
\item We continue-pretrain AdaPonderLM from PonderLM  up to 2.8B hyperparameters. We find that AdaPonderLM could maintain language modeling performance after continue pretrained by AdaPonderLM while decreasinig inference FLOPs.
\item We evaluate performance on downstream tasks, demonstrating capabilities competitive with existing models.
\item We conduct ablation studies on Pythia-70M to analyze the impact of hyperparameters $\lambda$ and $k$ (introduced in Section~\ref{sec:pretrain}), gating mechanisms, and settings introduced in Section~\ref{sec:mlp-gate-framework}.
\end{enumerate}
\subsection{Pretraining AdaPonderLM from Scratch}
\label{baselines}
We pretrain AdaPonderLM under the Pythia-70M and Pythia-410M architectures; detailed configurations are provided in~\autoref{appendix: model config}. All models are trained on 26B tokens sampled from the Pile~\citep{gao2020pile}. Unless otherwise specified, we set the target bottom-$K$ fraction to $k=0.1$ and the pondering regularization weight to $\lambda=0.1$.

To compare our model with other baselines, we also pretrained the following baselines, Loop Transformer~\citep{saunshi2025reasoning}, Pause Tokens~\citep{zelikman2024quiet}, and PonderLM~\citep{zeng2025ponderlm} as our baselines. All experiments are conducted with Pythia-70M and Pythia-410M suit and with approximately $4\times$ inference flops compared with vanilla Pythia backbone. Our specific hyperparameter settings are specifc in~\autoref{appendix: model config}.  Our result could be seen in~\autoref{tab:scaling_ppl}. 

\begin{table}[t]
\centering
\caption{Validation perplexity (PPL) and loss of baselines. }
\label{tab:scaling_ppl}
\vskip 0.1in
\begin{small}
\setlength{\tabcolsep}{4pt}
\begin{tabular}{lccccc}
\toprule
Backbone & \makecell{Inference FLOPs}& \#Params  & \makecell{Val Loss ($\downarrow$)} & \makecell{Val PPL ($\downarrow$)} \\
\midrule
\textbf{AdaPonderLM} & $3.8\times$ & 70M & \textbf{2.66}  & \textbf{14.32} \\
PonderLM & $4\times$ & 70M & 2.67 & 14.40\\  
Pause Token &$4\times$  &  70M & 2.79 & 16.27 \\
Loop Transformer &$4\times$  &  70M & 2.71 & 14.98 \\
\midrule
\textbf{AdaPonderLM} & $3.8\times$ & 410M & 2.29 & 9.87 \\
% PonderLM & $3\times$   & 410M & 2.28 & 9.78\\
PonderLM & $4\times$  & 410M & \textbf{2.27} & \textbf{9.72} \\
Pause Token &$4\times$  &  410M & 2.28 & 9.78 \\
Loop Transformer &$4\times$  &  410M & \textbf{2.27} & \textbf{9.72} \\
% MoR &$3.8\times$ &  410M &2.29 & 9.87 &   \\%%% 估算
\bottomrule
\end{tabular}
\end{small}
% \vskip -0.25in
\end{table}

We find that on the 70M backbone, \textbf{AdaPonderLM} achieves the best PPL among all recurrent baselines while using fewer inference FLOPs ($3.8\times$ vs. $4\times$), indicating that the learned token-wise halting can prune redundant recurrence without hurting modeling quality. On the 410M backbone, AdaPonderLM attains comparable validation loss/PPL to PonderLM and Loop Transformer, again with a ($\sim$)6--8\% reduction in inference compute. The results suggest AdaPonderLM exits early on easier tokens and spends more iterations on harder ones, yielding modest compute savings without degrading language modeling quality.
% We pretrain AdaPonderLM under the Pythia-70M and Pythia-410M architectures; detailed configurations are provided in Appendix~\ref{appendix: model config}. All models are trained on 25B tokens sampled from the Pile~\citep{gao2020pile}. 

% Unless otherwise specified, we set the target bottom-$K$ fraction to $k=0.1$ and the pondering regularization weight to $\lambda=0.1$ for the experiments in this section and the following one. 
% We further ablate these hyperparameters in Section~4.4.1 (Table~\ref{tab:ablation_hparams}), which indicates that the method is more sensitive to $\lambda$ while remaining relatively stable under reasonable choices of $k$. 
% All experiments are conducted with a two-stage schedule, using 4B tokens for Stage~1 (warm-up) and 6B tokens for Stage~2 (with pondering regularization).

% Furthermore, we continue pretrained PonderLM 1.4B and PonderLM 2.8B with 10B tokens, 5B tokens, respectively. We also continue pretrained vanilla Pythia-1.4B with 10 tokens. We warmed up the ponder gates with 1B tokens. The results are specific in Table~\ref{tab:scaling_ppl}.

\begin{table}[t]
\centering
\caption{Validation perplexity (PPL) and loss of AdaPonderLM and PonderLM under different Pythia backbones. }
\label{tab:continue pretrain}
\vskip 0.1in
\begin{small}
\setlength{\tabcolsep}{2pt}
\begin{tabular}{lccccc}
\toprule
Backbone & \makecell{Inference FLOPs}& \makecell{Training\#Tokens}& \#Params  & \makecell{Val Loss ($\downarrow$)} & \makecell{Val PPL ($\downarrow$)} \\
% \midrule
% \multicolumn{5}{l}{\textit{(Train from scratch)}} \\
% AdaPonderLM& $3.7\times$ & 25B & 70M   & \textbf{2.662} & \textbf{14.32} \\
% PonderLM & $3\times$ & 25B & 70M & 2.681 & 14.60 \\
% PonderLM & $4\times$ & 25B & 70M & 2.667 & 14.40\\
% \midrule
% AdaPonderLM& 3 & 25B & 410M  & 2.xx & xx.x \\
% PonderLM & $3\times$  & 25B & 410M & 2.281 & 9.78\\
% PonderLM & $4\times$  & 25B & 410M & 2.274 & 9.72 \\
\midrule
% \multicolumn{5}{l}{\textit{(Continue pretrain)}} \\
AdaPonderLM & $3.7\times$  & 310B & 1.4B & 1.92 & 6.82\\
% \multicolumn{5}{l}{\scriptsize{(CPT from PonderLM)}} \\
% AdaPonderLM & $3.7\times$  & 312B & 1.4B & 2.00 & 7.40 \\
% \multicolumn{5}{l}{\scriptsize{(CPT from Vanilla Model)}} \\
PonderLM & $4\times$  & 300B & 1.4B & 1.92 &  6.82\\
PonderLM & $4\times$  & 310B & 1.4B & 1.92 & 6.82   \\
% \multicolumn{5}{l}{\scriptsize{(CPT from PonderLM)}}\\
\midrule
AdaPonderLM & $3.8\times$  & 305B & 2.8B & 1.83 & 6.21 \\
% \multicolumn{5}{l}{\scriptsize{(CPT from PonderLM)}}\\
PonderLM & $4\times$  & 300B & 2.8B & 1.83 &  6.21\\
\bottomrule
\end{tabular}
\end{small}
% \vskip -0.3in
\end{table}

\subsection{Continue-Pretraining AdaPonderLM}

We also continue pretrain AdaPonderLM with PonderLM. We firstly freeze the Transformer backbone and warm up the initialized gate MLPs with 1B tokens before full fine-tuning.

As shown in \autoref{tab:continue pretrain}, AdaPonderLM maintains the modeling performance of the PonderLM baseline while reducing computational costs. Specifically, on both 1.4B and 2.8B scales, AdaPonderLM matches the validation loss of PonderLM (1.92 and 1.83) but \textbf{decreases the inference overhead} from $4\times$ to $3.7\times \sim 3.8\times$ FLOPs. This demonstrates that our method effectively accelerates recurrent models by identifying and pruning redundant iterations without compromising generation quality. 

\subsection{Downstream Task Evaluation}
\label{sec:downstream_eval}
 We use a set of widely adopted benchmarks, covering those originally employed in Pythia—LAMBADA~\citep{paperno2016lambada}, PIQA~\citep{bisk2020piqa}, WinoGrande~\citep{sakaguchi2021winogrande}, ARC-Easy and ARC-Challenge~\citep{clark2018think}, and SciQ~\citep{welbl2017crowdsourcing}. In addition, we evaluate commonsense reasoning on HellaSwag~\citep{zellers2019hellaswag} and reading comprehension on RACE~\citep{lai2017race}.  We compare our models against pretained models illustrated in Section~\ref{baselines} as well as large-scale pretrained language models such as OPT~\citep{zhang2022opt}, Bloom~\citep{muennighoff2023crosslingual}, Tinyllama~\citep{zhang2024tinyllama}, and GPTNeo~\citep{gpt-neo}.

Following the LM evaluation harness~\citep{eval-harness}, we report both zero-shot and five-shot performance. As shown in \autoref{tab:downstream}, AdaPonderLM outperforms vanilla Pythia baselines across 1.4B and 2.8B scales, achieving average accuracy gains of 2.2\% (zero-shot) and 3.5\% (five-shot) at the 2.8B level. Crucially, compared to the fixed-compute PonderLM, our method reduces inference FLOPs by approximately 10\% while maintaining highly competitive downstream performance. These results demonstrate that AdaPonderLM accelerates inference by pruning redundant computation without compromising generation quality.

\begingroup
\begin{table*}[t!]
\centering
\caption{Zero-shot and five-shot performance (accuracy, \%) on downstream tasks. For all baselines, we evaluate the released pretrained checkpoints from their official repositories. $\Delta$acc reports the mean accuracy improvement over the matched Pythia baseline.}
\label{tab:downstream}
% \vskip 0.1in
\begin{small}
\setlength{\tabcolsep}{2pt}   % 默认大约是 6pt，可以尝试 2~3pt
\begin{tabular}{@{}l|cccccccccc@{}}
% \begin{tabular}{@{}l@{\hspace{3pt}}|
% c@{\hspace{1.5pt}}c@{\hspace{1.5pt}}c@{\hspace{2pt}}c@{\hspace{2pt}}
% c@{\hspace{4pt}}c@{\hspace{4pt}}c@{\hspace{4pt}}c@{\hspace{4pt}}
% c@{\hspace{4pt}}c@{\hspace{3pt}}@{}}
\toprule
Model \scriptsize{(\#training tokens)} & \makecell{Lambada\\OpenAI} & \makecell{ARC \\ -E} & \makecell{Lambada\\Standard} & \makecell{ARC \\ -C} & \makecell{Wino \\ Grande} & PIQA & \makecell{Hella \\ Swag} & SciQ & RACE & \makecell{Avg acc /\\ $\Delta$acc $\uparrow$} \\

\midrule
\rowcolor{mygray}\multicolumn{11}{c}{\textbf{\texttt{{0-shot}}}}\\
% \midrule
% LLaMA-Loop1    & 53.6 & 54.3 & 41.8 & 23.6 & 52.9 & 69.3 & 35.8 & 83.6 & 33.5 & 49.8 \\
% LLaMA-Loop3    & 55.8 & 55.2 & 45.6 & 23.2 & 54.1 & 68.9 & 37.6 & 84.9 & 33.0 & 50.9 \\
% LLaMA-Pause1    & 46.0 & 55.6 & 36.9 & 23.9 & 51.1 & 69.2 & 35.7 & 90.1 & 27.9 & 48.5 \\
% LLaMA-Pause3    & 48.0 & 58.5 & 42.8 & 25.0 & 54.8 & 70.4 & 37.6 & 89.2 & 28.5 & 50.5 \\
% LLaMA-Ponder1    & 53.8 & 53.2 & 42.6 & 23.0 & 52.6 & 68.4 & 35.9 & 83.2 & 33.3 & 49.6 \\
% Ours    & 58.1 & 58.0 & 48.2 & 25.2 & 53.9 & 70.7 & 38.6 & 85.9 & 32.4 & 52.3 \\
% \midrule
% Latent-1B    & 61.8 & 67.3 & 58.4 & 33.3 & 61.1 & 75.1 & 48.5 & 90.7 & 37.3 & 59.3 \\
% Vanilla-1B    & 60.1 & 66.6 & 55.8 & 32.6 & 59.8 & 75.1 & 47.9 & 91.0 & 37.0 & 58.4 \\
% Latent-410M    & 59.1 & 54.0 & 47.3 & 24.6 & 55.5 & 69.4 & 37.7 & 86.2 & 33.5 & 51.9 \\
\midrule
Pythia-410M \scriptsize{(300B)}   & 51.4 & 52.2 & 36.4 & 21.4 & 53.8 & 66.9 & 33.7 & 81.5 & 30.9 & 47.6 \\
PonderLM-410M\scriptsize{(25B)}    & 45.8 & 48.7 & 33.5 & 20.3 & 51.5 & 65.9 & 32.4 & 79.7 & 29.9 & 45.3 \\
Loop Teansformer-410M \scriptsize{(25B)}& 44.3 & 48.1 & 31.9 & 20.0 & 51.3 & 64.1 & 32.1 & 79.3 & 31.2 & 44.7  \\
Pause Token-410M \scriptsize{(25B)} & 46.3 & 50.5 & 35.7 & 21.6 & 52.1 & 65.0 & 31.9 & 81.6 & 30.3 & 46.1 \\
\textbf{AdaPonderLM-410M} \scriptsize{(25B)}& 45.4 & 47.9 & 32.6 & 20.1 & 49.7 & 64.7 & 31.8 & 76.5 & 30.3  & 44.3 \\
\midrule
Pythia-1.4B \scriptsize{(300B)} & 61.6 & 60.4 & 49.7 & 25.9 & 57.5 & 70.8 & 40.4 & 86.4 & 34.1 & 54.1 \\
% \makecell[l]{\textbf{AdaPonderLM-1.4B} \scriptsize{(310B)}\\\scriptsize{(CPT from vanilla Pythia-1.4B)}}& 60.9 & 59.3 & 48.8 & 25.3 & 55.6 & 71.0 & 40.2 & 88.3 & 34.8 & 53.8 \\
OPT-1.3B \scriptsize{(300B)} & 57.9 & 57.1 & 52.5 & 23.4 & 59.7 & 71.8 & 41.6 & 84.3 & 34.3 & 53.6 \\
% GPTneo-1.3B \scriptsize{(300B)} & 57.1 & 56.2 & 45.3 & 23.2 & 55.0 & 71.2 & 38.6 & 86.1 & 34.5 & 51.9 \\
Bloom-1.7B \scriptsize{(366B)} & 46.2 & 56.4 & 44.5 & 23.7 & 56.8 & 68.5 & 37.5 & 85.0 & 33.2 & 50.2 \\
Ponder-1.4B \scriptsize{(300B)}& 65.2 & 62.0 & 53.8 & 27.0 & 60.1 & 72.6 & 44.0 & 89.0 & 35.2 & 56.5 / +2.4  \\
\textbf{AdaPonderLM-1.4B} \scriptsize{(312B)}& 64.5 & 61.5 & 52.4 & 27.1 & 58.0 & 72.2 & 43.7 & 88.7 & 34.6 & 55.9 / +1.8\\
\midrule
Tinyllama-1.1B \scriptsize{(3T)} & 58.8 & 60.3 & 49.3 &28.0& 59.0 & 73.3\textbf{ }& 45.0 & 88.9 & 36.4 & 55.4\textbf{ }\\
OPT-2.7B \scriptsize{(300B)} & 63.5 & 60.8 & 56.0 & 26.8 & 61.2 & 73.8 & 45.9 & 85.8 & 36.2 & 56.7 \\
GPTneo-2.7B \scriptsize{(300B)} & 62.1 & 61.2 & 51.6 & 27.5 & 57.8 & 72.3 & 42.7 & 89.3 & 35.1 & 55.5 \\
Bloom-3B \scriptsize{(366B)} & 51.7 & 59.4 & 50.9 & 28.0 & 58.7 & 70.8 & 41.4 & 88.8 & 35.2 & 53.9 \\
Pythia-2.8B \scriptsize{(300B)} & 64.6 & 64.4 & 54.3 & 29.5 & 60.2 & 73.8 & 45.4 & 88.5 & 34.9 & 57.3 \\
Pythia-6.9B \scriptsize{(300B)} & 67.2 & \textbf{67.3} & 55.9 & 31.4 & 61.0 & \textbf{75.2} & 48.1 & 89.3 & \textbf{36.9} & 59.1 \\
Ponder-2.8B \scriptsize{(300B)} & \textbf{68.9} & 66.5 & \textbf{60.8} & \textbf{32.5} & \textbf{63.6} & 75.0 & \textbf{48.6} & \textbf{91.0} & 36.5 & \textbf{60.4} / +3.1 \\
\textbf{AdaPonderLM-2.8B} \scriptsize{(305B)} & 68.3 & 65.3 & 59.8 & 31.4 & 62.0 & 75.0 & 48.6 & 90.8 & 36.3 & 59.6 / +2.2 \\
\midrule
\rowcolor{mygray}\multicolumn{11}{c}{\textbf{\texttt{5-shot}}}\\
\midrule
Pythia-410M \scriptsize{(300B)}   & 43.9 & 54.7 & 32.8 & 22.3 & 53.4 & 68.0 & 33.8 & 88.9 & 30.4 & 47.6 \\
PonderLM-410M\scriptsize{(25B)}    & 39.7 & 51.6 & 33.6 & 22.2 & 50.0 & 66.6 & 32.3 & 87.0 & 30.1 & 45.9 \\
Loop Teansformer-410M \scriptsize{(25B)}&  39.6 & 51.4 & 29.7 & 21.2 & 51.9 & 65.2 &  32.2 & 84.5 & 30.5 & 45.1 \\
Pause Token-410M \scriptsize{(25B)} & 40.3 & 51.7 & 32.8 & 21.8 & 50.5 & 65.6 & 31.8 & 87.4 & 29.0 & 45.7\\
\textbf{AdaPonderLM-410M} \scriptsize{(25B)}& 38.7 & 50.2 & 29.5 & 21.4 & 51.6 & 64.4 & 31.8 & 85.8 & 30.2 &  44.8  \\
% \textbf{Ours-410M (Pythia Arch)} \scriptsize{(300B)} & \textbf{52.1} & 58.0 & \textbf{45.0} & 26.0 & \textbf{54.6} & \textbf{69.2} & \textbf{37.9} & \textbf{91.7} & \textbf{32.6} & \textbf{51.9}
% / \textcolor{green!35!black}{+4.3} \\
\midrule
Pythia-1.4B \scriptsize{(300B)} & 54.5 & 63.1 & 44.5 & 28.8 & 57.1 & 71.0 & 40.5 & 92.4 & 34.6 & 54.1 \\
% \makecell[l]{\textbf{AdaPonderLM-1.4B} \scriptsize{(312B)}\\\scriptsize{(CPT from vanilla Pythia-1.4B)}}& 54.6 & 63.0 & 42.6 & 28.6 & 57.9 & 72.0 & 40.6 & 91.7 & 34.7 & 54.0\\
OPT-1.3B \scriptsize{(300B)}  & 54.0 & 60.4 & 49.0 & 26.9 & 56.9 & 72.4 & 38.5 & 91.8 & 35.4 & 52.7 \\
% GPTneo-1.3B \scriptsize{(300B)} & 49.9 & 59.9 & 44.5 & 25.9 & 56.6 & 71.6 & 38.6 & 86.1 & 34.5 & 51.9 \\
Bloom-1.7B \scriptsize{(366B)} & 42.5 & 58.8 & 41.5 & 26.2 & 57.7 & 68.7 & 37.6 & 91.9 & 33.5 & 50.9 \\
PonderLM-1.4B \scriptsize{(300B)} & 59.2 & 67.5 & 49.9 & 32.4 & 60.4 & 73.5 & 44.2 & 94.3 & 37.1 & 57.6 / +3.6  \\
\textbf{AdaPonderLM-1.4B} \scriptsize{(310B)} & 59.0 & 66.0 & 48.1 & 30.9 & 59.4 & 73.2 & 44.0 &94.0 & 36.3 & 56.8 / +2.8 \\
\midrule
Tinyllama-1.1B \scriptsize{(3T)} & 53.8 & 64.8 & 45.0 & 31.1 & 59.4 & 73.8 & 44.9 & 94.0 & 36.4 & 55.9 \\ % Empty cells follow reordering
OPT-2.7B \scriptsize{(300B)} & 60.2 & 64.7 & 55.0 & 29.8 & 62.2 & 75.1 & 46.1 & 93.0 & 37.5 & 58.2 \\
GPTneo-2.7B \scriptsize{(300B)} & 56.0 & 64.0 & 51.6 & 30.1 & 59.6 & 73.9 & 42.4 & 93.3 & 35.5 & 56.3 \\
Bloom-3B \scriptsize{(366B)} & 46.2 & 63.8 & 47.1 & 31.7 & 57.8 & 70.8 & 41.4 & 93.4 & 34.6 & 54.1 \\
Pythia-2.8B \scriptsize{(300B)} & 59.0 & 67.0 & 50.7 & 31.0 & 61.1 & 74.4 & 45.3 & 93.7 & 35.9 & 57.6 \\
Pythia-6.9B \scriptsize{(300B)} & 62.5 & 69.6 & 54.8 & 35.6 & 62.9 & 76.6 & 48.0 & 94.6 & 36.7 & 60.1 \\
PonderLM-2.8B \scriptsize{(300B)} & \textbf{64.2} & 70.6 & \textbf{58.7} & \textbf{35.8} & \textbf{65.3} & \textbf{76.7} & \textbf{49.0} & 94.3 & \textbf{39.0} & \textbf{61.5} / +3.9 \\
\textbf{AdaPonderLM-2.8B} \scriptsize{(300B)} & 63.5 & \textbf{70.9} & 58.0 & 35.2 & 63.6 & 75.9 & 48.9 & \textbf{95.0} & 38.5 & 61.1 / +3.5 \\
% OPT-2.7B \scriptsize{(300B)} & 63.5 & 60.8 & 56.0 & 26.8 & 61.2 & 73.8 & 45.9 & 85.8 & 36.2 & 56.7 \\
% GPTneo-2.7B \scriptsize{(300B)} & 62.1 & 61.2 & 51.6 & 27.5 & 57.8 & 72.3 & 42.7 & 89.3 & 35.1 & 55.5 \\
% Bloom-3B \scriptsize{(366B)} & 51.7 & 59.4 & 50.9 & 28.0 & 58.7 & 70.8 & 41.4 & 88.8 & 35.2 & 53.9 \\

% Add your LatentPythia-1B results here
\bottomrule
\end{tabular}
\end{small}
% \vskip -0.2in
\end{table*}
\endgroup
\subsection{Ablation Study}
We conduct ablation studies using the Pythia-70M architecture to analyze the impact of key hyperparameter and inference settings. All models are trained on a subset of the Pile dataset.

\subsubsection{How $k$ and $\lambda$ affects the result}
\begin{figure*}[t]
    % 第一个盒子放图片，占用 0.75 到 0.8 的宽度
    \begin{minipage}{0.7\linewidth}
\includegraphics[width=\linewidth]{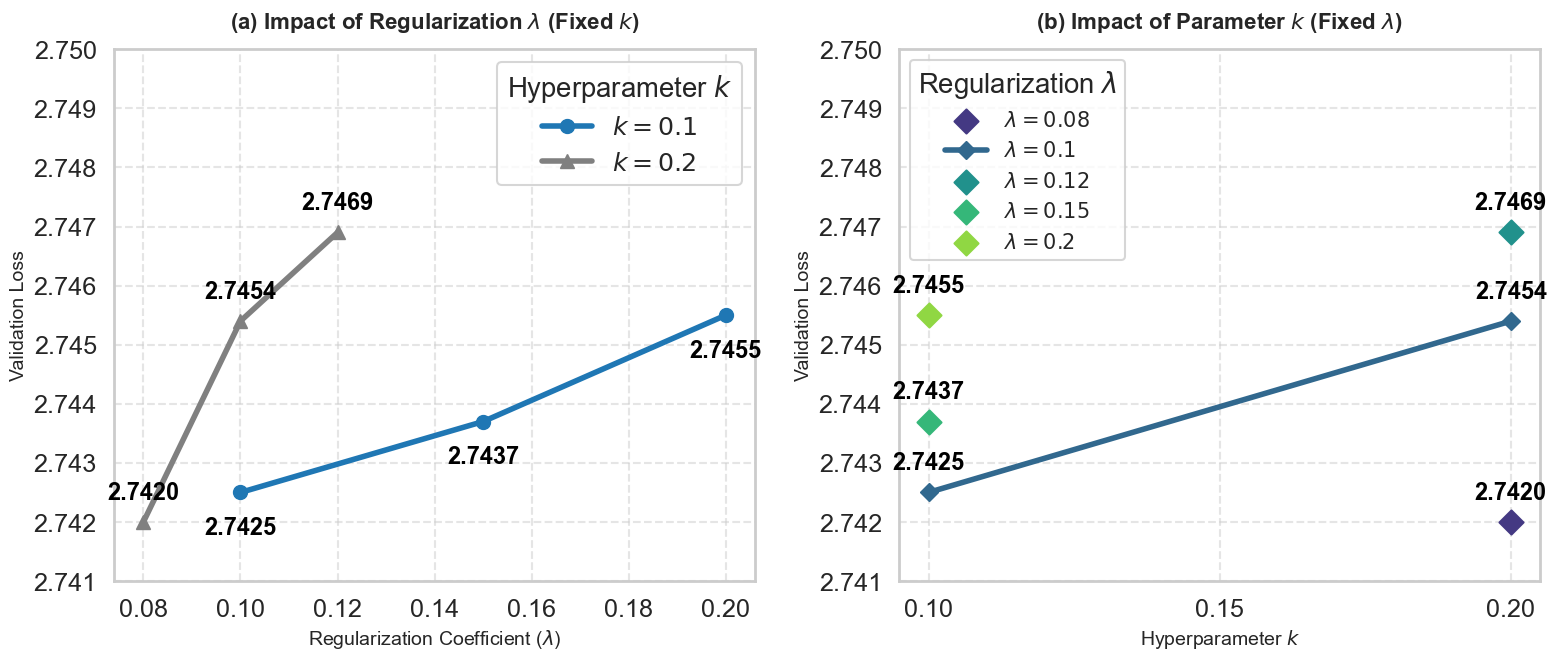}
    \end{minipage}% 注意这里不要空行，用 % 紧接着
    \hfill % 填充中间的空隙
    % 第二个盒子放标题，占用剩余宽度（稍微留点余量防止换行溢出）
    \begin{minipage}{0.28\linewidth}
        % 使用 \captionof 需要 caption 包，ICML 默认通常支持，
        % 如果报错 undefined \captionof，请在导言区加 \usepackage{caption}
        \caption{Hyperparameter Sensitivity. Loss increases monotonically with $\lambda$ (2.7425 $\rightarrow$ 2.7455) and $k$ (2.7425 $\rightarrow$ 2.7454), indicating that lower regularization strengths are preferred for minimizing loss.}
        \label{fig:ablation_para}
    \end{minipage}
    % \vskip -0.2in
\end{figure*}
We evaluate the target bottom-$K$ fraction $k$ within the set $\{0.1, 0.15, 0.2\}$ and vary the pondering penalty weight $\lambda$ across the interval $[0.08, 0.20]$ (specifically testing points such as $0.08, 0.1, 0.12,$ and $0.15$). We purpose our result in~\autoref{fig:ablation_para}.

We analyze the impact of regularization coefficient $\lambda$ and pruning fraction $k$. Results show a monotonic degradation in loss as regularization strengthens. At fixed $k=0.1$, increasing $\lambda$ from 0.1 to 0.2 raises the loss from 2.7425 to 2.7455. Similarly, with fixed $\lambda=0.1$, doubling $k$ from 0.1 to 0.2 increases loss to 2.7454. 
The evaluation loss is highly sensitive to these hyperparameters, where smaller values of $\lambda$ and $k$ consistently yield superior performance.

Although we observe a monotonic increase in validation loss as the regularization strength $\lambda$ and target pruning fraction $k$ increase, these parameters serve as essential control knobs for the efficiency-performance frontier. In the absence of strict regularization (i.e., lower $\lambda$ or $k$), the gating mechanism lacks the incentive to halt early, often resulting in a negligible pruning rate. Consequently, calibrating $\lambda$ and $k$ is necessary to push the model away from the 'safe' strategy of full-depth recurrence and towards an adaptive regime that balances modeling fidelity with computational sparsity.

\subsubsection{Impact of the gate mechanism}
\begin{figure}[t]
\centering
\includegraphics[width=0.8\linewidth]{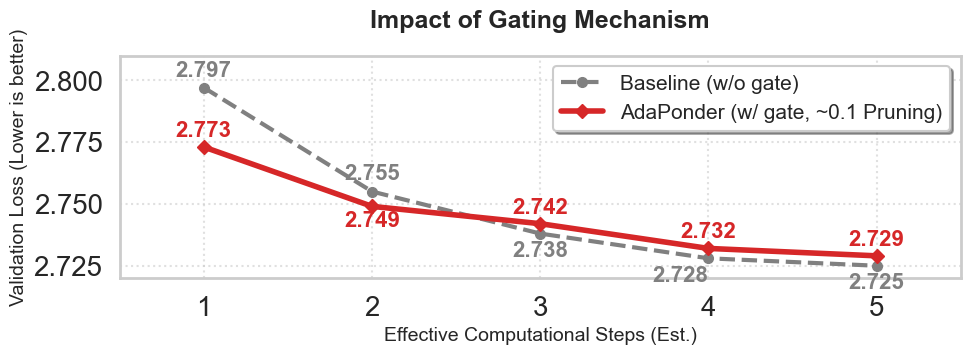}
\caption{Gating mechanism reduces effective compute steps by 10\% without compromising representation quality.}
\label{fig:ablation_prune}
% \vskip -0.25in
\end{figure}
To evaluate adaptive efficiency, we compare AdaPonderLM with a 10\% prune rate and varying ponder steps. We benchmark AdaPonder against the dense baseline across varying iterations. As shown in~\autoref{fig:ablation_prune}, at 5 iterations, the baseline achieves a loss of 2.725. AdaPonder attains a comparable loss of 2.729 with $\sim$10\% token pruning,  outperforming the 3-step baseline (2.738) and matching the 4-step performance (2.728).

\subsubsection{Insufficiency for a single MLP gate}

%% 我们在70M模型上尝试了使用单个MLP作为预测方法。实验发现，如果使用单个MLP，模型非常容易发生全部都不halt或者所有token都在第一步就停止halt了。这意味着使用单个MLP很难训练稳定

%% 同时，我们发现，如果不使用bottomK算法，模型很容易陷入所有token都不prune或者所有token都prune的情形。这表明bottomK算法是必要的
We investigate a simplified gating design in which a single MLP is used to directly predict the halting decision for each token in the  70M-parameter model. Empirically, we find that such a design is highly unstable during training. In particular, the model frequently collapses to degenerate solutions, where either almost no tokens halt across all iterations, or all tokens halt prematurely at the first step. This behavior indicates that a single MLP gate lacks sufficient inductive bias to produce a well-calibrated and stable halting policy. 

Moreover, we observe that removing the $\text{bottom}K$ selection mechanism further exacerbates this collapse. In the absence of $\text{bottom}K$, the model is prone to trivial equilibria in which either all tokens are pruned or none are pruned at each step. 

\section{Analysis: How MLPs Decide the Pruning Steps}
\label{sec:interpretability}
\begin{wrapfigure}{r}{0.5\linewidth}
    \centering
    \includegraphics[width=\linewidth]{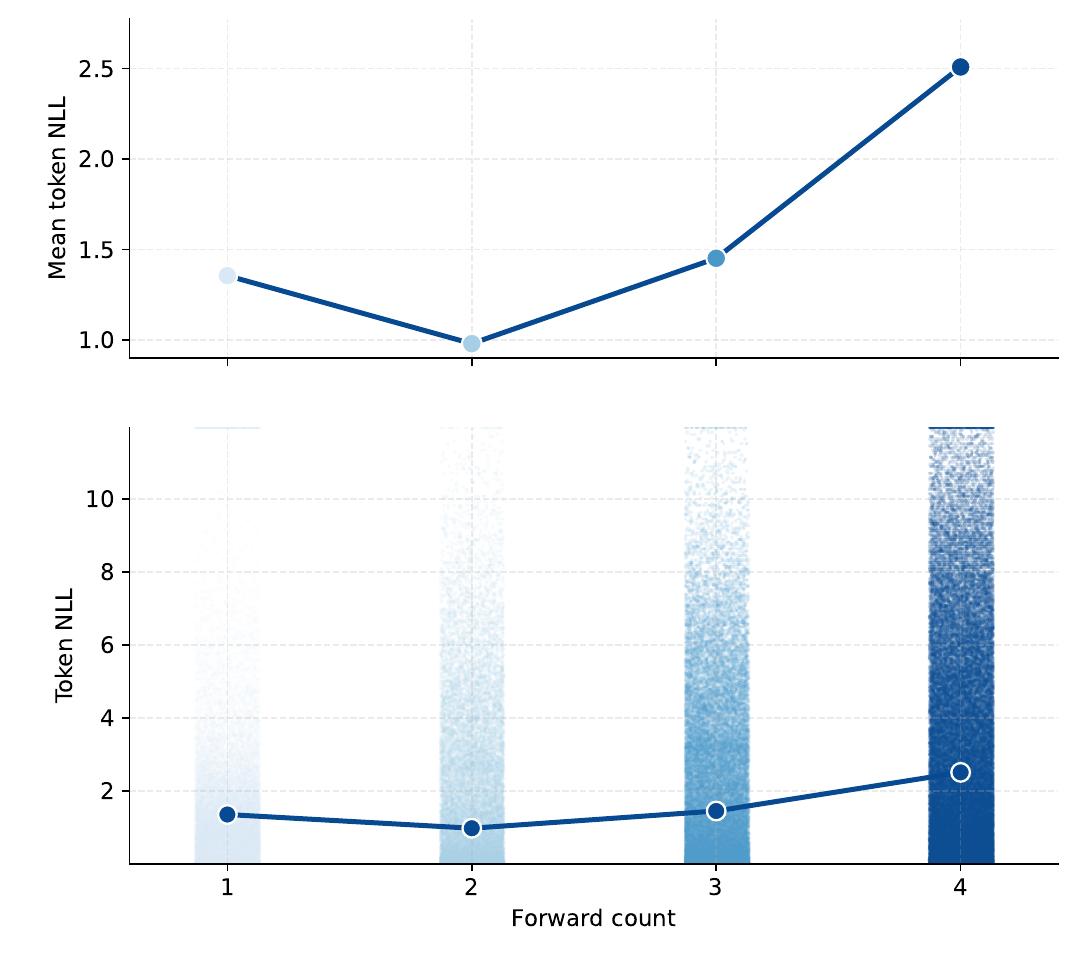}
    \caption{\textbf{Dynamics of pruning across recurrent iterations.} \textbf{Top}: Mean token NLL for tokens halted at different forward steps. \textbf{Bottom}: Density map of the token NLL distribution (clipped for clarity).}
    \label{fig:convergence_dynamics}
\vskip -0.6in
\end{wrapfigure}  
To better understand the decision-making of the learned MLP gates, we study how token difficulty correlates with the computational budget allocated at inference. Specifically, we analyze the pretrained \textbf{AdaPonder-410M} model on a subset of the Pile validation set. We focus on two questions: (i) whether the model follows the Adaptive Computation Time (ACT) principle by spending fewer recurrent steps on ``easy'' tokens while reserving deeper computation for ``hard'' tokens; and (ii) whether the learned gating mechanism within PonderLM is near-optimal, or if a predefined adaptive policy achieves comparable results.
\subsection{Convergence Dynamics of Pruned vs. Unpruned Tokens}
\label{sec:convergence_dynamics}
% 为了了解被prune token的相关性质，我们统计了验证集里每一轮被prune的token和这些token的平均NLL，见Figure~\ref{fig:convergence_dynamics}

To better understand the characteristics of pruned tokens, we collect, on the validation set, the tokens pruned at each iteration and compute their mean NLL, as shown in ~\autoref{fig:convergence_dynamics}.

% \begin{figure*}[t]

  % \vskip 0.3in
% \begin{minipage}{0.48\linewidth}
%     \includegraphics[width=\linewidth]{vertical_loss_comparison.pdf}
%     \caption{\textbf{Adaptive Policy vs. Fixed Distributions.} \textbf{Top}: Eval loss on the Pile under Iso-FLOPs. \textbf{Bottom}: Effective keep probability per step for each strategy.}
%     \label{fig:pile_comparison}
%     % \vspace{-0.8in} % 调整底部间距
% \end{minipage}

% \end{figure*}

\begin{figure}
    \centering
    \includegraphics[width=\linewidth]{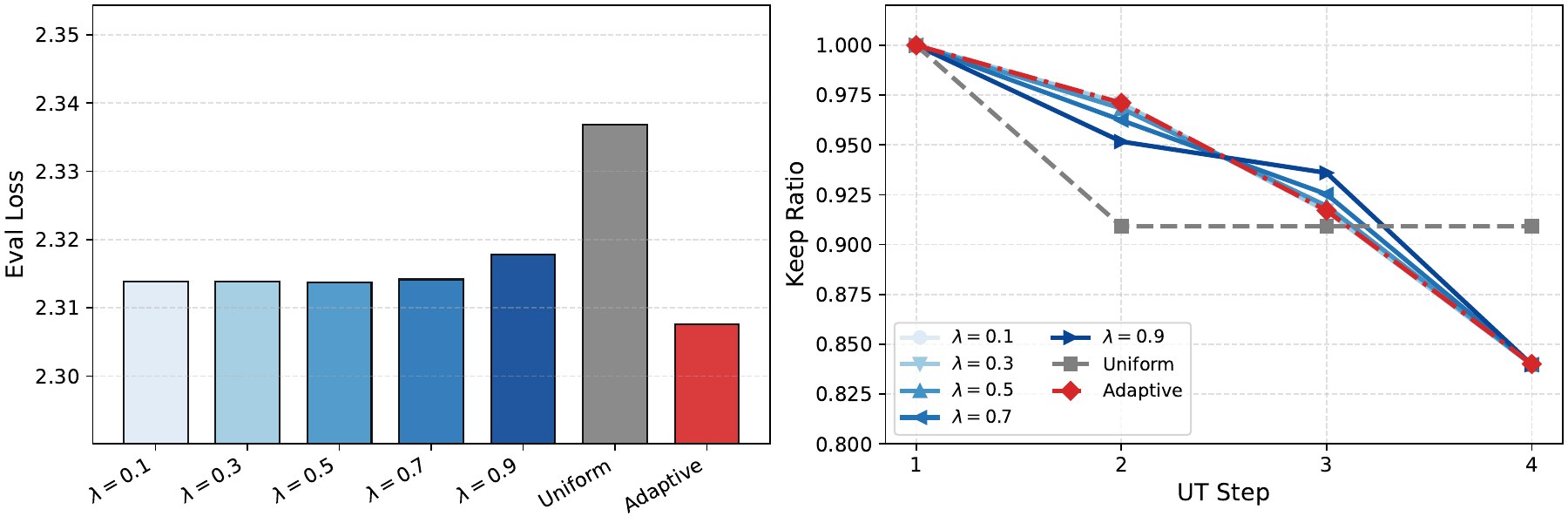}
    \caption{\textbf{Adaptive Policy vs. Fixed Distributions.} \textbf{Top}: Eval loss on the Pile under Iso-FLOPs. \textbf{Bottom}: Effective keep probability per step for each strategy.}
    \label{fig:pile_comparison}
\end{figure}
 The trend is non-monotonic: the mean NLL follows a ``check-mark'' shape, with the lowest value at Step~2, while later steps increasingly concentrate high-uncertainty tokens.

% \begin{wrapfigure}{r}{0.6\linewidth}
%     \centering

%     % \vskip -0.3in
% \end{wrapfigure}
\begin{itemize}
    \item \textbf{Confidence consolidation (Step 2).} Tokens halted at Step~2 achieve the minimum mean NLL ($\approx 1.0$), suggesting a brief refinement stage that resolves most unambiguous tokens after an initial screening.
    \item \textbf{Hard-token accumulation (Steps 3--4).} Mean NLL rises thereafter, peaking at Step~4 ($\approx 2.5$). The density map is heavy-tailed, indicating that later iterations disproportionately retain semantically ambiguous or long-range dependent tokens that need more computation.
\end{itemize}
% 移除那些强行断页的注释和命令，直接开始 wrapfigure

% 用 \noindent 和 \textbullet 伪造列表项
% \noindent\textbullet\quad \

Overall, the halting policy behaves as a staged filter rather than a simple difficulty threshold: early pruning, high-confidence resolution at Step~2, and progressively more compute for harder tokens. This suggests a strong link between a token's difficulty and the inference depth it receives.

\subsection{Comparison with Fixed-Distribution Strategies}
\label{sec:dist_comparison}

\textbf{Setup.} We compare the learned adaptive policy against two fixed baselines: (i) \textbf{Uniform}, which retains tokens with a constant keep probability across iterations; and (ii) \textbf{Geometric}, where the keep probability decays geometrically with parameter $\lambda$. Using the pretrained AdaPonderLM-410M, we impose a target pruning ratio at each iteration by pruning the $k$ tokens with the smallest gate values at that step. All strategies are evaluated under an \textbf{Iso-FLOP} constraint, i.e., with matched average numbers of active tokens, on a subset of the Pile validation set.

\textbf{Results.} As shown in~\autoref{fig:pile_comparison}, the Adaptive policy achieves the best eval loss under the same compute budget. Uniform performs worst, and Geometric variants remain consistently below Adaptive, indicating that gains come from token-specific, dynamic depth allocation rather than a static pruning distribution. It reveals that our learning strategy could guide LM to find the best propotion in gating mechaism.
\section{Limitations, and Future Work}
%% 我们的最大优势：我们提出了一种可以自主调节forward次数的递归神经网络方法。这种方法无需任何特殊的预先设定，模型即可自主学习出最优比例的剪枝方式。同时，我们提出了一套
%% 缺点：一定的参数敏感。为了保证模型性能且能进行一定的剪枝，lambda和k的设置不能过于突出
%% 未来工作：由于本工作发现了NLL和模型剪枝的关系。未来工作可能会通过不同层间NLL的差值作为监督信号。

There are three limitations in this work. First, to balance language modeling performance and pruning ratio, our method is sensitive to the choice of \(k\) and \(\lambda\) during pretraining. Second, our experiments are restricted to relatively small language models and datasets; the scalability of our approach to full-scale pretraining has not been thoroughly validated. Finally, introducing the MLP, although only slightly increasing the parameter count, still adds additional trainable parameters and training overhead.

We outline two future directions. First, we will continue scaling our approach to larger models and datasets. Second, motivated by the strong correlation we observe between token-wise NLL and pruning decisions, we plan to incorporate a supervised signal directly into the token-wise NLL, rather than relying on an MLP for interpretation.

\section{Conclusion}
In this work, we introduced \textbf{AdaPonderLM}, an end-to-end self-supervised framework that addresses the inefficiency of fixed recurrence steps in recurrent language models. Building on a Ponder-style refinement process, AdaPonderLM employs iteration-specific MLP gates and a monotonic persistent mask to dynamically determine the optimal recurrence depth for each token, eliminating the need for external supervision or manually tuned pruning ratios. Crucially, we proposed a KV-reuse mechanism to ensure consistency between training and inference, enabling practical acceleration by reusing cached states for halted tokens.

We validated \textbf{AdaPonderLM} across Pythia backbones ranging from 70M to 2.8B parameters, including both pretraining from scratch and continued pretraining settings. Our results demonstrate that AdaPonderLM reduces inference FLOPs by approximately 10\% while matching or exceeding the language modeling performance and downstream task accuracy of fixed-depth baselines like PonderLM. Furthermore, our analysis confirms that the learned gating policy aligns with the principles of Adaptive Computation Time, automatically allocating deeper recurrence to tokens with higher intrinsic difficulty. These findings suggest that self-supervised adaptive recurrence is a promising direction for efficient test-time scaling.
\bibliography{example_paper}
% \bibliographystyle{icml2026}

%%%%%%%%%%%%%%%%%%%%%%%%%%%%%%%%%%%%%%%%%%%%%%%%%%%%%%%%%%%%%%%%%%%%%%%%%%%%%%%
%%%%%%%%%%%%%%%%%%%%%%%%%%%%%%%%%%%%%%%%%%%%%%%%%%%%%%%%%%%%%%%%%%%%%%%%%%%%%%%
% APPENDIX
%%%%%%%%%%%%%%%%%%%%%%%%%%%%%%%%%%%%%%%%%%%%%%%%%%%%%%%%%%%%%%%%%%%%%%%%%%%%%%%
%%%%%%%%%%%%%%%%%%%%%%%%%%%%%%%%%%%%%%%%%%%%%%%%%%%%%%%%%%%%%%%%%%%%%%%%%%%%%%%
\newpage
\appendix
\onecolumn

\section{The Model Configure of The Main Experiment}
\label{appendix: model config}
We specific our main hyperparameter for all pretraining models in~\autoref{table:hyper-wiki}.

\begin{table}[ht]
\caption{The hyperparameter for the pretrained model.}
\label{table:hyper-wiki}
\vskip 0.15in
\begin{center}
\begin{small}
\begin{sc}
\begin{tabular}{lr|lr|lrlr}
\toprule
Hyperparameter & Value & Hyperparameter & Value & Hyperparameter & Value \\
\midrule
Batch Size & 256 & Max Positional Embedding & 2048 & ZerO Stage &0\\
Dropout & 0   & Weight Decay & 0.1 & BF16 & True \\
Warmup Ratio   & 0.02  & MLP Hidden States & $4h$ & lr scheduler & cosine \\
Adam Beta1   & 0.9  & Adam Beta 2 & 0.95& Adam Epsilon & 1e-8 \\

\bottomrule
\end{tabular}
\end{sc}
\end{small}
\end{center}
\vskip -0.2in
\end{table}

During the pretraining phase, we adopted EleutherAI’s learning rate configuration. Specifically, the learning rates were set to $1\times10^{-3}$ for the 70M model, $3\times10^{-4}$ for the 410M model,  $2\times10^{-4}$ for the 1.4B model, and $1.2\times10^{-4}$ for the 2.8B model. For the continued pretraining phase, the learning rate was scaled to one-tenth of the value used in pretraining.

We specify the detailed training configurations as follows:
\begin{itemize}
\item \textbf{AdaPonderLM Pretraining:} We set Stage 1 to 20,000 steps and Stage 2 to 30,000 steps, with a warmup schedule starting at $S_0=20000$ and ending at $S_1=24,000$.
\item \textbf{Baseline Alignment:} For the Pause Token model, we append three pause tokens after each training token to align inference FLOPs. For MoR, we set the keeping ratios $k=[0.97, 0.93, 0.9]$ across layers to approximate the pruning ratio of AdaPonderLM.
\item \textbf{Continue-Pretraining (CPT):} For CPT from \textbf{PonderLM}, we perform joint training of the MLP gates and backbone after an initial gate-only training phase, with $S_1=4,000$. 
\end{itemize}

\end{document}